\documentclass[sigconf]{acmart} 

\usepackage{microtype}
\usepackage{inconsolata}
\usepackage{amsmath}
\usepackage[ruled,vlined]{algorithm2e}
\usepackage{graphicx}
\usepackage{booktabs}
\usepackage{makecell}
\usepackage{multirow}
\usepackage{array}
\usepackage{tabularx}
\usepackage[table]{xcolor}
\usepackage{tcolorbox}
\usepackage{caption} 
\usepackage{lipsum}  
\usepackage{amsmath}
\newcommand{\uparr}{\ensuremath{\uparrow}}

\tcbset{colback=gray!5!white, colframe=black, width=\columnwidth}
\usepackage{float}        
\newfontfamily\bangla[Script=Bengali]{kalpurush.ttf}

\definecolor{ash}{gray}{0.4} 
\AtBeginDocument{%
  }

\setcopyright{acmlicensed}
\copyrightyear{2018}
\acmYear{2018}
\acmDOI{XXXXXXX.XXXXXXX}
\acmConference[Conference acronym 'XX]{Make sure to enter the correct
  conference title from your rights confirmation email}{June 03--05,
  2018}{Woodstock, NY}
\acmISBN{978-1-4503-XXXX-X/2018/06}

\begin{document}

\title{When Less Is Enough: Context Selection and Prompting Strategies for Bengali News Headline Generation}

\author{Muhammad Ashad Kabir}
\authornote{Corresponding author.}
\affiliation{%
  \institution{Charles Sturt University}
  \city{Bathurst}
  \state{NSW}
  \country{Australia}
}
\email{akabir@csu.edu.au}

\author{Kawsar Ahmed}
\affiliation{%
  \institution{Chittagong University of Engineering and Technology}
  \city{Chittagong}
  \country{Bangladesh}}

\author{Md. Osama}
\affiliation{%
  \institution{Bangladesh Army University of Science and Technology}
  \city{Nilphamari}
  \country{Bangladesh}
}

\renewcommand{\shortauthors}{Kabir et al.}

\begin{abstract}
 Large language models (LLMs) have shown strong performance in text generation tasks, yet their effectiveness on headline generation remains sensitive to how input context is selected and presented. In this work, we investigate Bengali news headline generation as a document-level generation task that requires effective selection and presentation of salient contextual information from long-form articles. Using Gemini-2.0-Flash, Llama-3.3-70B, and GPT-4o, we systematically study the effects of context selection, prompting strategies, and in-context learning (i.e., few-shot) on the quality of headline generation. Our experiments show that providing the full article does not necessarily improve performance; instead, using selected lead paragraphs of the article can maintain, and in some cases improve, headline generation quality. We further compare Bengali Native Prompting (BNaP) and Cross-Lingual Prompting (XLP), and examine how each interacts with context-enriched prompt templates incorporating auxiliary contextual cues. Results demonstrate that prompting strategies substantially influence generation quality: XLP often yields stronger performance, particularly when combined with contextual enrichment, but its benefits are model-dependent. Additionally, few-shot prompting substantially improves Gemini, with most of the gain obtained from a single demonstration, whereas Llama shows limited benefit from additional examples.
 Overall, our findings highlight that effective Bengali news headline generation depends more on context relevance and prompt design than on increasing input length, offering practical insights for multilingual and low-resource LLM applications.
\end{abstract}




\maketitle

\section{Introduction}

Automatic news headline generation is a central task in abstractive summarization, aiming to condense long-form articles into concise, informative, and engaging titles. Early approaches primarily relied on encoder-decoder architectures and transformer-based models such as BART~\cite{lewis2019bart}, T5~\cite{raffel2020exploring}, and BanglaT5~\cite{bhattacharjee-etal-2023-banglanlg}. More recently, large language models (LLMs) have demonstrated remarkable capabilities in headline and summary generation, often outperforming traditional fine-tuned models in zero- and few-shot settings~\cite{ding2023harnessing}. Instruction-tuned and demonstration-based prompting methods have further improved LLM performance in text generation tasks~\cite{jiang-etal-2024-instruction, kojima2022large}.

Despite these advances, LLMs are fundamentally constrained by their input context window. When processing long documents, they can lose essential information, and longer prompts may even degrade performance due to loss of focus, semantic dilution, or capacity overflow~\cite{liu2023lost}. Moreover, \citet{chhabra-etal-2024-revisiting} extend this discussion to the broader notion of position bias, showing that both LLMs and pretrained summarization models tend to disproportionately prioritize information from certain parts of the input text. This observation has motivated research on source–summary alignment and context utilization in summarization, where summary content is mapped to corresponding portions of the source document to examine which input positions contribute to the generated summary \cite{fan2018controllable,ravaut2024context}. Furthermore, incorporating additional contextual information, such as sentiment, aspect, and category, has been shown to improve headline generation robustness~\cite{OSAMA2025100138}. However, in low-resource languages like Bengali, the use of LLMs for headline generation remains relatively underexplored, making this an important research direction~\cite{akash-etal-2023-shironaam}.

In this paper, we investigate how selective context conditioning and prompting strategies affect LLM-based Bengali news headline generation. 
To identify the most informative portions of news articles, we first analyze the semantic similarity between gold headlines and individual article paragraphs using sentence-transformer-based representations. Our analysis shows that the opening paragraphs consistently exhibit the highest semantic alignment with the target headlines, suggesting that salient contextual information is concentrated early in the articles. Motivated by this observation, we introduce a selective context conditioning strategy in which LLMs are prompted using only the first one through five paragraphs of an article instead of the full document. This setup enables us to systematically examine whether relevance-focused context selection can improve headline generation quality while reducing unnecessary contextual noise.

We further investigate two prompting paradigms: Bengali Native Prompting (\textit{BNaP}) and Cross-Lingual Prompting (\textit{XLP}). For each paradigm, we evaluate both a \textit{Baseline} prompt containing only the news content and enriched \textit{MultiGen} prompt templates that incorporate additional contextual cues, including sentiment, aspect, category, and definition-based information. Additionally, we compare zero-shot and few-shot settings to examine how in-context demonstrations influence prompt effectiveness across different prompting strategies.
Our key findings are as follows:
\begin{itemize}
    \item We show that selectively conditioning LLMs on articles' lead paragraphs can match, and for some models outperform, full-document conditioning for Bengali headline generation, highlighting the importance of context relevance over input length.
    \item We demonstrate that in-context learning (few-shot prompting) can enhance performance, while different LLMs exhibit varying sensitivities to the number of demonstrations.
    \item We find that prompting strategies strongly influence performance: Cross-Lingual Prompting (\textit{XLP}) often benefits Gemini and Llama, especially under contextual enrichment, while prompt effectiveness remains model-dependent. 
\end{itemize}

\section{Related Work}

\subsection{News Summarization and Headline Generation}

Early work on news headline generation leveraged transformer-based encoder-decoder architectures such as BART~\cite{lewis2019bart} and T5~\cite{raffel2020exploring}, which achieved strong performance in abstractive summarization tasks. These pretrained models were subsequently benchmarked for both headline generation and news summarization, demonstrating their effectiveness across a variety of datasets. More recently, LLMs have been shown to excel in zero-shot and few-shot headline generation tasks, especially when instruction-tuned or prompted with demonstrations~\cite{ ding2023harnessing, jiang-etal-2024-instruction, kojima2022large}. Despite these advances, prior work has largely overlooked how context selection and prompt strategies jointly affect headline generation quality in low-resource settings such as Bengali.

\subsection{Prompting Strategies and Contextual Enrichment}
Prompt design plays a central role in shaping LLM generation quality, particularly in zero-shot and few-shot settings. Instruction-tuned LLMs have demonstrated strong task generalization capabilities through demonstration-based prompting~\cite{jiang-etal-2024-instruction}, while prompting strategies and cross-lingual prompting have shown that prompt structure and language can substantially influence model behavior~\cite{kojima2022large, qin2023cross}. Recent work has further explored enriching prompts with auxiliary contextual information, such as sentiment, category, and aspect metadata, to improve generation relevance and controllability in news-related tasks~\cite{OSAMA2025100138,akash-etal-2023-shironaam}. However, excessively long or redundant prompts may dilute attention over salient information and degrade generation quality~\cite{liu2023lost}. Despite these advances, the interaction between multilingual prompting strategies, prompting enriched with contextual cues, and selective context conditioning remains underexplored for low-resource headline generation tasks such as Bengali news headline generation.

\subsection{Context Selection}

Although modern LLMs support increasingly large context windows, generation quality often degrades when models process long and information-dense inputs. \citet{beltagy2020longformer} introduce sparse local and global attention mechanisms that scale linearly with sequence length, enabling efficient processing of substantially longer documents. \citet{liu2023lost} show that the performance of LLMs can degrade substantially when relevant information appears at different positions within long input contexts, indicating that these models do not consistently utilize information across the entire context. This finding underscores the importance of identifying and selecting the most relevant contextual segments for effective long-context processing. Prior summarization research has identified position bias as a common characteristic of both LLMs and pretrained summarization models, where information from certain positions in the input is disproportionately favored during summary generation~\cite{chhabra-etal-2024-revisiting}. This phenomenon is particularly relevant to news summarization, where the placement of salient information can influence generation quality. However, the role of selective context conditioning in LLM-based headline generation remains under-explored, particularly for low-resource languages such as Bengali.

\begin{figure*}[!ht]
    \centering
    \includegraphics[width=1\textwidth]{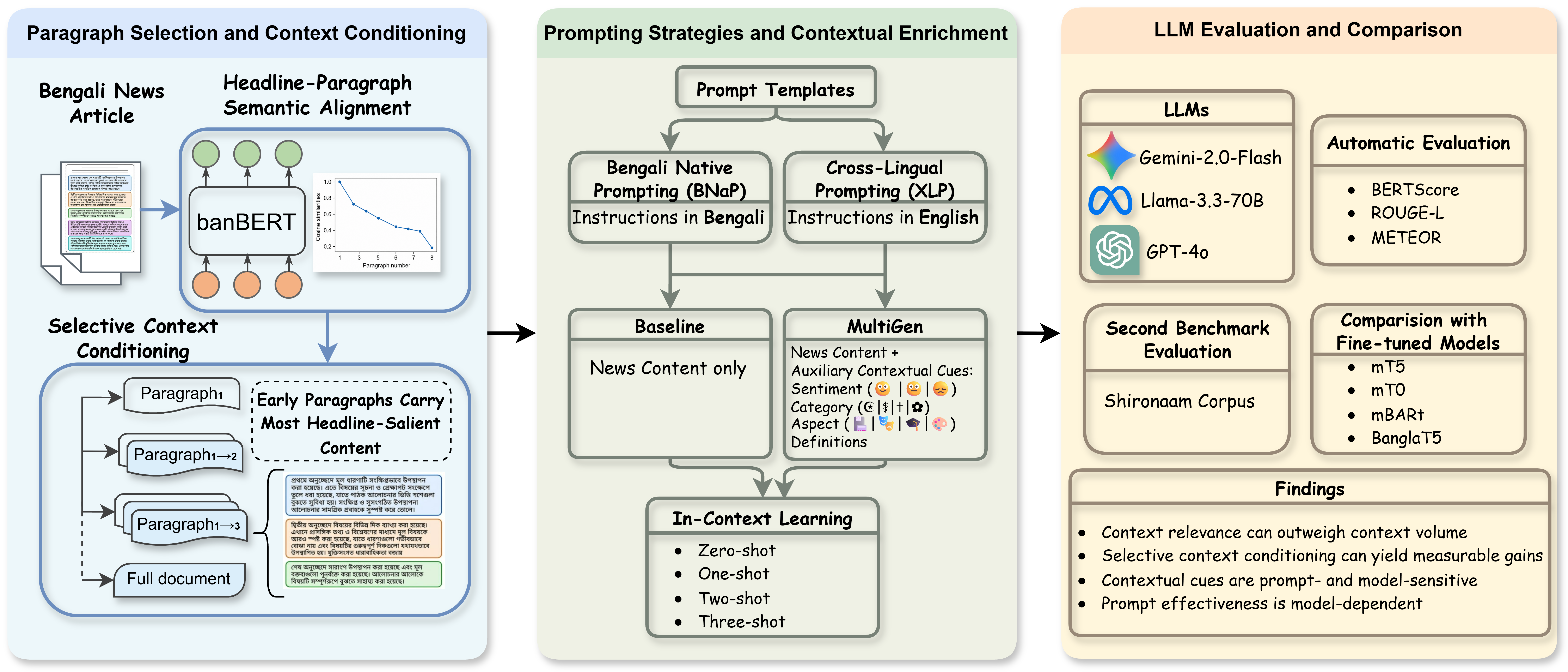}
    \caption{Overview of our headline-generation workflow, illustrating paragraph selection, context conditioning, prompting strategies, contextual enrichment, and LLM evaluation.}
    \label{fig:method-diagram}
\end{figure*}

\subsection{Low-Resource and Multilingual Headline Generation}

Low-resource languages such as Bengali remain under-explored in headline generation and abstractive summarization. While multilingual datasets such as XL-Sum \cite{hasan2021xl} have facilitated summarization research across multiple languages, recent work has shown that cross-lingual retrieval-augmented in-context learning can substantially improve zero-shot performance on Bangla NLP tasks by leveraging semantically similar examples from high-resource languages, demonstrating the effectiveness of cross-lingual prompting for low-resource settings \cite{li-etal-2023-crosslingual}. Recent multilingual headline generation studies have further demonstrated that pretrained transformer-based encoder-decoder models can effectively transfer knowledge across languages, improving generation quality even in low-resource settings through multilingual pretraining and cross-lingual learning \cite{litvak-etal-2019-ranlp, bhattacharjee-etal-2023-banglanlg}. More recently, \citet{akash-etal-2023-shironaam} and \citet{OSAMA2025100138} investigated Bengali headline generation using auxiliary contextual features and fine-tuned multilingual encoder-decoder models. However, the role of LLM-based prompting strategies, selective context conditioning, and in-context learning remains largely unexplored for Bengali headline generation.

Among prior Bengali headline-generation studies, \citep{akash-etal-2023-shironaam} and \citep{OSAMA2025100138} are particularly relevant to our work because they combine auxiliary information with the full content of the news articles for headline generation. Our study addresses a different but complementary question. Rather than performing content-specific sentence retrieval or introducing a new context-selection algorithm, we investigate whether a simple position-based lead-context strategy is sufficient for general-purpose LLMs, and systematically examine how the amount of article context interacts with prompt language, contextual enrichment, and in-context demonstrations. Thus, our contribution lies primarily in characterizing these interactions for LLM-based Bengali headline generation and identifying when reduced lead context can preserve or improve generation quality while substantially reducing input length.

\section{Methodology}
An overview of our headline generation methodology is illustrated in Figure~\ref{fig:method-diagram}.

\subsection{Datasets}
We employ two benchmark datasets in this study. As the primary dataset, we use \textit{BeliN}~\cite{OSAMA2025100138}, a curated Bengali religious news corpus comprising 2,520 articles collected from multiple national newspapers. Each record contains six fields: \textit{Source}, \textit{Headline}, \textit{Article}, \textit{Category} (religious affiliation), \textit{Aspect} (thematic focus), and \textit{Sentiment}. 

As a secondary benchmark, we use the \textit{Shironaam} dataset~\cite{akash-etal-2023-shironaam}, which contains diverse Bengali news articles along with auxiliary contextual information such as \textit{category}, \textit{topic words}, and \textit{image captions}.

\subsection{Semantic Relevance Between Headlines and Paragraphs}
To estimate the semantic relevance between headlines and article content, we compute cosine similarity between headline embeddings and paragraph embeddings using the \texttt{banBERT}\footnote{\url{https://huggingface.co/banglagov/banBERT-Base}} sentence representation model. 
For each article, the paragraph achieving the highest cosine similarity with the reference headline is identified as the most semantically aligned paragraph. The complete procedure is summarized in Algorithm~\ref{alg:headline_similarity} (Appendix~\ref{append:a}). Importantly, this analysis is performed on the BeliN training set and uses reference headlines only to characterize where headline-relevant information tends to occur within articles; the reference headline is not used to select context for test-time headline generation.
This analysis enables us to examine how salient contextual information is distributed across article paragraphs and provides empirical motivation for our selective context conditioning strategy.

\subsection{Selective Context Conditioning}
To investigate how context selection affects LLM-based Bengali headline generation, we condition the model on only the initial paragraphs of a news article rather than the full document. This strategy is motivated by the journalistic observation that lead paragraphs often contain the most salient information and is further supported by our semantic similarity analysis, which shows that early paragraphs exhibit the strongest alignment with reference headlines.

Formally, let a news article $A$ consist of paragraphs $\{p_1, p_2, ..., p_n\}$. We define a partial context as:
\[
paragraph_{1 \rightarrow k} = \{p_1, p_2, ..., p_k\},
\]
where $k \in \{1,2,3,\dots\}$ denotes the number of initial paragraphs provided to the model. By varying $k$, we examine how different amounts of contextual information influence headline generation quality and whether relevance-focused partial context can outperform full-document conditioning.

\subsection{Prompting Strategies}

We investigate the impact of linguistic framing and contextual enrichment on Bengali news headline generation using two prompting paradigms: \textit{Bengali Native Prompting (BNaP)} and \textit{Cross-Lingual Prompting (XLP)}. \textit{BNaP} expresses the instruction entirely in Bengali, following prior evaluations of Bengali-language prompting~\cite{ahmed-etal-2025-bennumeval}.
In contrast, \textit{XLP} uses English instructions while retaining Bengali news articles as input, leveraging the cross-lingual capabilities of multilingual LLMs \cite{qin2023cross}.

For each prompting paradigm, we evaluate two prompt variants. The \textit{Baseline} variant includes only the news article content, whereas \textit{MultiGen} enriches the prompt with auxiliary contextual information, including concise definitions and metadata such as \textit{Category}, \textit{Aspect}, and \textit{Sentiment}. Combining these settings yields four prompting templates: \textit{BNaP+Baseline}, \textit{BNaP+MultiGen}, \textit{XLP+Baseline}, and \textit{XLP+MultiGen} (Tables~\ref{tab:prompt_templates_xlp_baseline}--\ref{tab:prompt_templates_xlp_multigen} in Appendix~\ref{Prompt_templates}). 

To further examine the effect of in-context learning, we evaluate \textit{zero-shot}, \textit{one-shot}, \textit{two-shot}, and \textit{three-shot} configurations, where each demonstration consists of a complete article-headline example embedded within the prompt.

\subsection{Large Language Models}
We evaluate Gemini-2.0-Flash \cite{googledeepmind2025gemini20flash} (hereafter referred to as ``Gemini 2.0"), Llama-3.3-70B (hereafter referred to as ``Llama-3.3") \cite{meta2024llama33}, and GPT-4o~\cite{openai2024gpt4o}, selected to represent proprietary and open-source multilingual LLMs with strong instruction-following capabilities. Gemini-2.0-Flash and Llama-3.3-70B are used in the context-scope and in-context learning experiments, while GPT-4o is additionally included in the prompting-strategy comparison to broaden the model-family analysis. 

To examine the effects of prompting and contextual enrichment, we evaluate each model using \textit{BNaP} and \textit{XLP} prompting paradigms under both \textit{Baseline} and \textit{MultiGen} configurations. We further compare \textit{zero-shot}, \textit{one-shot}, \textit{two-shot}, and \textit{three-shot} settings to analyze the influence of in-context demonstrations on headline generation quality.

\subsection{Evaluation}
We evaluate generated Bengali headlines using three widely adopted metrics: \texttt{BERTScore}~\cite{zhang2020bertscoreevaluatingtextgeneration}, \texttt{METEOR}~\cite{banerjee2005meteor}, and \texttt{ROUGE-L}~\cite{lin2004rouge}. These metrics jointly capture semantic similarity, lexical overlap, and structural alignment between generated and reference headlines. For all reported metrics, we present the mean score across test instances together with the standard deviation (SD) and 95\% bias-corrected and accelerated (BCa) bootstrap confidence intervals (CIs) computed using 10,000 resamples. Key pairwise comparisons are summarized using paired mean differences with 95\% BCa bootstrap confidence intervals. Statistical significance of paired instance-level differences is assessed using two-sided Wilcoxon signed-rank tests. These statistics quantify variation across test instances for the fixed generation outputs obtained in each experimental condition.

To ensure consistency across experimental conditions, all LLM experiments use the same decoding settings: temperature = 0.7, top-$p$ = 0.9, and a maximum output length of 50 tokens. For few-shot prompting, demonstrations are selected by random sampling from the training set and then used consistently across models and prompting strategies for each shot setting, with the same demonstration examples and ordering. Our goal is therefore not to optimize demonstration retrieval but to isolate the effects of context selection, prompt language, contextual enrichment, and the number of in-context demonstrations. For input-length analysis, we estimate token counts using the \texttt{o200k\_base}\footnote{\url{https://github.com/openai/tiktoken}} tokenizer as a consistent tokenizer proxy across settings.

We use the official \textit{BeliN} test set (505 articles) for all main headline generation experiments to ensure fair comparison with prior work. We also evaluate on the \textit{Shironaam} test set (15,012 articles) for the second-benchmark comparison.

\section{Results and Discussion}
\label{sec:result}
\subsection{Impact of Selective Context Conditioning}
\label{sec:res:sel-context}

\subsubsection{Semantic Alignment Between Headlines and Article Paragraphs}
To examine how salient information is distributed across news articles, we compute semantic similarity between reference headlines and individual article paragraphs using \texttt{banBERT} embeddings on 2,015 training articles from the \textit{BeliN} dataset. Our analysis reveals a strong positional bias: the first paragraph exhibits the highest similarity to the reference headline in 56.48\% of articles, as shown in Table~\ref{tab:similarity_distribution}. This observation aligns with the journalistic ``inverted pyramid'' structure, where key information is typically concentrated at the beginning of an article \cite{po2003news}.

\begin{table}[!t]
\centering
\caption{Distribution of paragraph positions with the semantic similarity to the reference headline, showing frequency and cumulative percentage across the \textit{BeliN} training set.}
\label{tab:similarity_distribution}
\begin{tabular}{lrc}
\toprule
Paragraph & Articles, \textit{n} (\%) & Cumulative, \textit{n} (\%) \\
\midrule
1st    & 1138 (56.48)  & 1138 (56.48)  \\
2nd    & 238 (11.81)   & 1376 (68.29)  \\
3rd    & 155 (7.69)    & 1531 (75.98)  \\
4th    & 104 (5.16)    & 1635 (81.14)  \\
5th    & 74 (3.67)     & 1709 (84.81)  \\
6th    & 82 (4.07)     & 1791 (88.88)  \\
7th    & 56 (2.78)     & 1847 (91.66)  \\
$\ge$8th & 168 (8.34) & 2015 (100.00) \\
\bottomrule
\end{tabular}
\end{table}

The cumulative distribution further motivates the use of selective context conditioning. The first two paragraphs contain the highest-similarity content for 68.29\% of articles, while the first three paragraphs account for 75.98\%. Beyond the third paragraph, the likelihood of observing the most semantically aligned content decreases substantially, indicating diminishing contextual relevance in later sections of the article.

\begin{figure}[!t]
\centering
\includegraphics[width=0.48\textwidth]{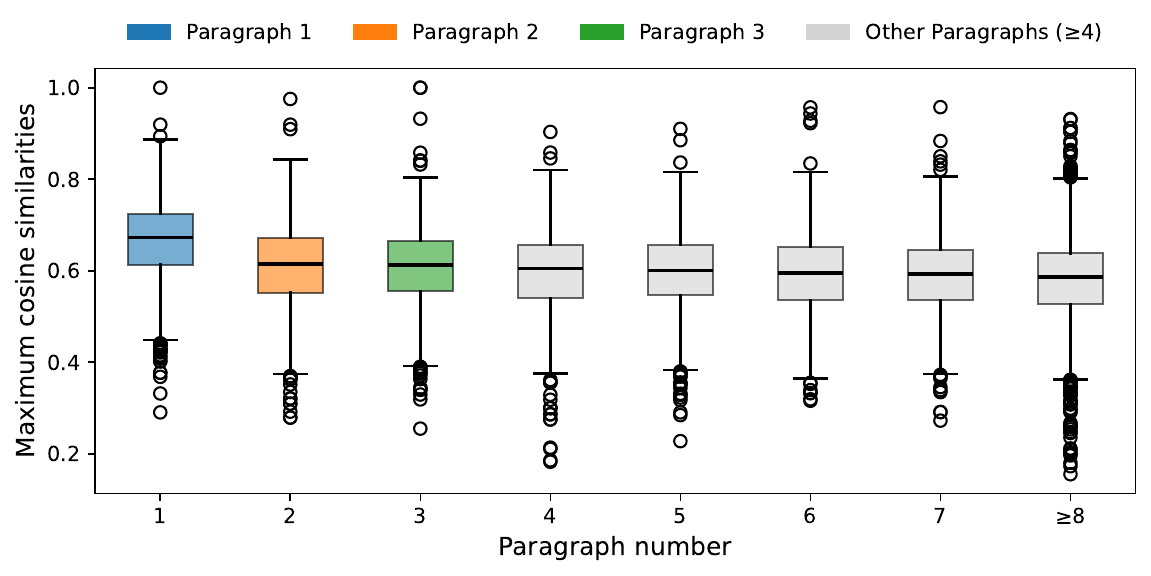}
\caption{Paragraph-wise distribution of cosine similarity scores between reference headlines and article content, illustrating the positional concentration of salient headline-related information in early paragraphs.}
\label{fig:similarity_scores}
\end{figure}

Figure~\ref{fig:similarity_scores} further supports this trend by illustrating the distribution of cosine similarity scores across paragraph positions. The first paragraph not only exhibits the highest frequency of alignment with the reference headline but also achieves the highest median similarity score. We observe a consistent decline in similarity from Paragraph 1 to Paragraph 3, after which the distributions stabilize at lower levels. Although a small subset of articles exhibits salient information in later paragraphs, the overall results suggest that the initial paragraphs often contain the content most semantically aligned with the reference headline, supporting the use of selective context conditioning for LLM-based headline generation.

\subsubsection{Contextual Salience vs. Information Volume}
To examine the relationship between contextual salience and input information volume, we evaluate headline generation quality under varying selective context conditioning settings using the \textit{BNaP+Baseline} prompting configuration. The resulting performance across different context scopes is summarized in Table~\ref{tab:result_full_vs_partial_updated}. 

\begin{table*}[!t]
\centering
\caption{Performance of Gemini and Llama across varying context scopes (partial vs. full article input), illustrating the effect of selective context conditioning for the \textit{BNaP+Baseline} prompt template. Values denote mean $\pm$ SD, and brackets indicate 95\% BCa bootstrap CI estimated over test instances.}
\begin{tabular}{c c c c c c c}
\toprule
Article & \multicolumn{3}{c}{Gemini-2.0-Flash} & \multicolumn{3}{c}{Llama-3.3-70B} \\
\cmidrule(lr){2-4} \cmidrule(lr){5-7}
 & BERTScore \uparr & ROUGE-L \uparr & METEOR \uparr & BERTScore \uparr & ROUGE-L \uparr & METEOR \uparr \\
\midrule
\makecell[t]{$Full$} &  \makecell[t]{$0.719_{\color{ash}\pm0.054}$\\{\small\color{ash}[0.715, 0.724]}} & \makecell[t]{$0.199_{\color{ash}\pm0.146}$\\{\color{ash}\small[0.187, 0.212]}} & \makecell[t]{$0.224_{\color{ash}\pm0.198}$\\{\color{ash}\small[0.207, 0.242]}} 
& \makecell[t]{$0.754_{\color{ash}\pm 0.064}$\\{\color{ash}\small[0.749, 0.760]}} & \makecell[t]{$0.243_{\color{ash}\pm 0.194}$\\{\color{ash}\small[0.227, 0.261]}} & \makecell[t]{$0.164_{\color{ash}\pm 0.203}$\\{\color{ash}\small[0.147, 0.183]}} \\

\makecell[t]{$paragraph_{1}$} & \makecell[t]{$\textbf{0.727}_{\color{ash}\pm0.064}$\\{\color{ash}\small[0.722, 0.732]}} & \makecell[t]{$0.209_{\color{ash}\pm0.182}$\\{\color{ash}\small[0.194, 0.225]}} & \makecell[t]{$0.207_{\color{ash}\pm0.210}$\\{\color{ash}\small[0.189, 0.226]}} & \makecell[t]{$0.747_{\color{ash}\pm 0.063}$\\{\color{ash}\small[0.742, 0.753]}} & \makecell[t]{$0.230_{\color{ash}\pm 0.201}$\\{\color{ash}\small[0.213, 0.248]}} & \makecell[t]{$0.154_{\color{ash}\pm 0.194}$\\{\color{ash}\small[0.138, 0.172]}}\\

\makecell[t]{$paragraph_{1\rightarrow{2}}$} & \makecell[t]{$0.721_{\color{ash}\pm 0.057}$\\{\color{ash}\small[0.716, 0.726]}} & \makecell[t]{$0.204_{\color{ash}\pm 0.155}$\\{\color{ash}\small[0.191, 0.218]}} & \makecell[t]{$0.208_{\color{ash}\pm 0.193}$\\{\color{ash}\small[0.192, 0.225]}} & \makecell[t]{$0.750_{\color{ash}\pm 0.064}$\\{\color{ash}\small[0.744, 0.756]}} & \makecell[t]{$0.237_{\color{ash}\pm 0.194}$\\{\color{ash}\small[0.221, 0.255]}} & \makecell[t]{$0.159_{\color{ash}\pm 0.194}$\\{\color{ash}\small[0.144, 0.178]}} \\

\makecell[t]{$\textbf{\textit{paragraph}}_{1\rightarrow{3}}$} & \makecell[t]{$0.720_{\color{ash}\pm 0.056}$\\{\color{ash}\small[0.715, 0.725]}} & \makecell[t]{$0.204_{\color{ash}\pm 0.153}$\\{\color{ash}\small[0.191, 0.217]}} & \makecell[t]{$0.213_{\color{ash}\pm 0.199}$\\{\color{ash}\small[0.197, 0.231]}} & \makecell[t]{$\textbf{0.755}_{\color{ash}\pm 0.064}$\\{\color{ash}\small[0.749, 0.761]}} & \makecell[t]{$\textbf{0.243}_{\color{ash}\pm 0.191}$\\{\color{ash}\small[0.227, 0.261]}} &	\makecell[t]{$\textbf{0.170}_{\color{ash}\pm 0.197}$\\{\color{ash}\small[0.153, 0.188]}}\\

\makecell[t]{$paragraph_{1\rightarrow{4}}$} & \makecell[t]{$0.721_{\color{ash}\pm 0.058}$\\{\color{ash}\small[0.716, 0.727]}} & \makecell[t]{$0.206_{\color{ash}\pm 0.156}$\\{\color{ash}\small[0.193, 0.220]}} & \makecell[t]{$0.226_{\color{ash}\pm 0.206}$\\{\color{ash}\small[0.209, 0.245]}} & 	\makecell[t]{$0.752_{\color{ash}\pm 0.061}$\\{\color{ash}\small[0.746, 0.757]}} & \makecell[t]{$0.241_{\color{ash}\pm 0.189}$\\{\color{ash}\small[0.225, 0.257]}} & \makecell[t]{$0.158_{\color{ash}\pm 0.185}$\\{\color{ash}\small[0.143, 0.175]}} \\
$paragraph_{1\rightarrow{5}}$ & \makecell[t]{$0.720_{\color{ash}\pm 0.058}$\\{\color{ash}\small[0.714, 0.724]}} & \makecell[t]{$0.200_{\color{ash}\pm0.153}$\\{\color{ash}\small[0.189, 0.211]}} & \makecell[t]{$0.214_{\color{ash}\pm 0.201}$\\{\color{ash}\small[0.197, 0.231]}} & \makecell[t]{$0.752_{\color{ash}\pm 0.064}$\\{\color{ash}\small[0.747, 0.758]}} & \makecell[t]{$0.243_{\color{ash}\pm 0.197}$\\{\color{ash}\small[0.227, 0.261]}} &	\makecell[t]{$0.164_{\color{ash}\pm 0.199}$\\{\color{ash}\small[0.148, 0.182]}}\\
\bottomrule
\end{tabular}
\label{tab:result_full_vs_partial_updated}
\end{table*}

\begin{table}[t]
\centering
\caption{Input-token reduction achieved by first-three-paragraph selective context conditioning on the test set.}
\label{tab:token_reduction}
\resizebox{\linewidth}{!}{
\begin{tabular}{lcccc}
\toprule
\textit{BeliN} test set & Count ($n$) & \multicolumn{3}{c}{Input tokens}\\
\cmidrule(lr){3-5}
& & Full article & First 3 paragraphs & Reduction \\
\midrule
All articles & 504 & $1011.60_{\pm665.89}$ & $479.35_{\pm398.29}$ & 52.6\% \\
$>3$ paragraphs & 378  & $1089.44_{\pm683.51}$ & $379.82_{\pm267.93}$ & 65.1\% \\
\bottomrule
\end{tabular}
}
\end{table}

A notable observation is that both Gemini-2.0-Flash and Llama-3.3-70B achieve their strongest performance using partial article context rather than the full document. Gemini performs best with only the first paragraph ($paragraph_{1}$), achieving a \texttt{BERTScore} of 0.727 (95\% CI: 0.722--0.732). Compared with the full-article input, this corresponds to a statistically significant mean paired improvement of 0.0078 \texttt{BERTScore} points (95\% CI: 0.003--0.012; $p=0.0011$). Llama attains its highest observed \texttt{BERTScore} when using the first three paragraphs ($paragraph_{1 \rightarrow 3}$), with a score of 0.755 (95\% CI: 0.749--0.761). Although the estimated difference is positive ($\Delta$ \texttt{BERTScore} = 0.0008), this effect is not statistically significant (95\% CI: $-0.0071$ to $0.0088$, $p = 0.82$), indicating no measurable performance difference between the reduced-context and full-context conditions.

In addition to preserving headline generation quality, selective context conditioning substantially reduces input length (Table~\ref{tab:token_reduction}). On the full \textit{BeliN} test set, using the first three paragraphs reduces the average input length from 1011.60$\pm$665.89 tokens to 479.35$\pm$398.29 tokens, corresponding to a 52.6\% reduction. Since 126 test articles contain three or fewer paragraphs, for which the first-three-paragraph input is effectively identical to the full document, we further examine articles with more than three paragraphs. For this subset, the average input length decreases from 1089.44$\pm$683.51 tokens to 379.82$\pm$267.93 tokens, corresponding to a 65.1\% reduction. These results show that the first-three-paragraph setting provides substantial input reduction while maintaining comparable performance for Llama; Gemini’s best-performing first-paragraph setting suggests that even shorter salient contexts can be beneficial for some models.

Taken together, these results suggest that headline generation quality depends more on contextual relevance and model-specific context utilization than on simply increasing the amount of input text. For Gemini, conditioning on concise lead content yields measurable gains, whereas Llama remains largely robust to additional context. While longer inputs provide broader coverage, they may also introduce redundant or less relevant information that dilutes attention to task-relevant content. This observation aligns with prior work showing that LLM performance can degrade in long and information-dense contexts due to reduced utilization of salient information \cite{liu2023lost}. 

\subsubsection{Model-Specific Sensitivity to Contextual Salience}
The two models exhibit notably different sensitivities to contextual salience within the input article. Llama maintains relatively stable performance from $paragraph_{1 \rightarrow 3}$ (0.755) to the full article (0.754), differing by only 0.001 \texttt{BERTScore} points (Table~\ref{tab:result_full_vs_partial_updated}). In contrast, Gemini performs best using only the shortest context ($paragraph_{1}$) and shows no consistent improvement from additional input paragraphs.
These results suggest that LLMs differ in their ability to utilize longer and information-dense contexts for headline generation. While Llama appears more tolerant of additional contextual content, Gemini benefits more from concise and highly salient inputs. Overall, our findings indicate that increasing context length does not necessarily improve generation quality and that lead-context conditioning can provide an effective alternative to full-document input.

\subsection{Effect of In-Context Learning}
Motivated by the findings in Section~\ref{sec:res:sel-context}, which show that the first three paragraphs capture the majority of semantically salient information, we evaluate the impact of in-context learning (ICL) on headline generation using the \textit{BNaP+Baseline} prompting configuration with partial article input ($paragraph_{1 \rightarrow 3}$). Table~\ref{tab:result_few_shot} presents results for Gemini and Llama under zero-shot, one-shot, two-shot, and three-shot settings, highlighting model-specific differences in responsiveness to demonstration-based prompting.

\begin{table*}[t!]
\centering
\caption{Impact of in-context learning (ICL) across zero-shot to three-shot settings for Gemini and Llama using the \textit{BNaP+Baseline} prompting configuration with partial article input ($paragraph_{1 \rightarrow 3}$).
}
\label{tab:result_few_shot}
\begin{tabular}{c c c c c c c}
\toprule
Shot & \multicolumn{3}{c}{Gemini-2.0-Flash} & \multicolumn{3}{c}{Llama-3.3-70B} \\
\cmidrule(lr){2-4} \cmidrule(lr){5-7}
 & BERTScore \uparr & ROUGE-L \uparr & METEOR \uparr & BERTScore \uparr & ROUGE-L \uparr & METEOR \uparr\\
\midrule
\textit{Zero-shot} & \makecell[t]{$0.720_{\color{ash}\pm 0.056}$\\{\color{ash}\small[0.715, 0.725]}} & \makecell[t]{$0.204_{\color{ash}\pm 0.153}$\\{\color{ash}\small[0.191, 0.217]}} & \makecell[t]{$0.213_{\color{ash}\pm 0.199}$\\{\color{ash}\small[0.197, 0.231]}} & \makecell[t]{$0.755_{\color{ash}\pm 0.064}$\\{\color{ash}\small[0.749, 0.761]}} & \makecell[t]{$0.243_{\color{ash}\pm 0.191}$\\{\color{ash}\small[0.227, 0.261]}} &\makecell[t]{$0.170_{\color{ash}\pm 0.197}$\\{\color{ash}\small[0.153, 0.188]}}\\
\textit{One-shot} & \makecell[t]{$0.760_{\color{ash}\pm 0.068}$\\{\color{ash}\small[0.754, 0.766]}} & \makecell[t]{$0.281_{\color{ash}\pm 0.212}$\\{\color{ash}\small[0.263, 0.300]}} & \makecell[t]{$0.245_{\color{ash}\pm 0.243}$\\{\color{ash}\small[0.224, 0.267]}} & \makecell[t]{$\textbf{0.761}_{\color{ash}\pm 0.068}$\\{\color{ash}\small[0.755, 0.767]}} & \makecell[t]{$\textbf{0.257}_{\color{ash}\pm 0.209}$\\{\color{ash}\small[0.240, 0.276]}} & \makecell[t]{$\textbf{0.175}_{\color{ash}\pm 0.208}$\\{\color{ash}\small[0.158, 0.194]}}\\
\textit{Two-shot} & \makecell[t]{$0.759_{\color{ash}\pm 0.069}$\\{\color{ash}\small[0.753, 0.765]}} & \makecell[t]{$0.283_{\color{ash}\pm 0.209}$\\{\color{ash}\small[0.266, 0.301]}} & \makecell[t]{$0.241_{\color{ash}\pm 0.235}$\\{\color{ash}\small[0.221, 0.262]}} & \makecell[t]{$0.756_{\color{ash}\pm 0.065}$\\{\color{ash}\small[0.750, 0.762]}} & \makecell[t]{$0.251_{\color{ash}\pm 0.203}$\\{\color{ash}\small[0.234, 0.269]}} & \makecell[t]{$0.163_{\color{ash}\pm 0.199}$\\{\color{ash}\small[0.147, 0.182]}}\\
\textit{Three-shot} & 	\makecell[t]{$\textbf{0.760}_{\color{ash}\pm 0.069}$\\{\color{ash}\small[0.754, 0.766]}} & \makecell[t]{$\textbf{0.283}_{\color{ash}\pm 0.210}$\\{\color{ash}\small[0.266, 0.302]}} & \makecell[t]{$\textbf{0.251}_{\color{ash}\pm 0.246}$\\{\color{ash}\small[0.230, 0.273]}} & \makecell[t]{$0.754_{\color{ash}\pm 0.062}$\\{\color{ash}\small[0.749, 0.760]}} & \makecell[t]{$0.250_{\color{ash}\pm 0.195}$\\{\color{ash}\small[0.234, 0.268]}} & \makecell[t]{$0.164_{\color{ash}\pm 0.187}$\\{\color{ash}\small[0.149, 0.181]}}\\
\bottomrule
\end{tabular}
\end{table*}

\subsubsection{Performance Gains}
Gemini improves substantially when in-context demonstrations are introduced, with performance largely plateauing after one-shot prompting. Its \texttt{BERTScore} rises from 0.72 (95\% CI: 0.715--0.725) in the zero-shot setting to 0.76 (95\% CI: 0.754--0.766) with three-shot prompting (Table~\ref{tab:result_few_shot}). Relative to the zero-shot baseline, this corresponds to a statistically significant mean paired improvement of 0.04 \texttt{BERTScore} points (95\% CI: 0.032--0.047; $p<0.001$). \texttt{ROUGE-L} and \texttt{METEOR} also remain above the zero-shot setting, indicating that few-shot prompting improves overall generation quality for Gemini.

In contrast, Llama shows comparatively stable performance across few-shot settings. Its highest \texttt{BERTScore} of 0.761 (95\% CI: 0.755--0.767) is achieved with one-shot prompting, followed by slight declines in the two-shot and three-shot configurations. 
However, the improvement from the zero-shot to the one-shot setting is not statistically significant ($\Delta$ \texttt{BERTScore} = 0.0056, 95\% CI: $-0.0024$ to 0.0138; $p=0.252$), indicating
that Llama is less dependent on multiple demonstrations and can generate strong outputs with limited in-context guidance.

\subsubsection{Model-Specific Sensitivity}
The contrasting trends indicate that the effectiveness of ICL is model-dependent, but neither model consistently benefits from increasing the number of demonstrations beyond one. Gemini shows a substantial improvement when moving from zero-shot to one-shot prompting, after which performance (ROUGE-L and METEOR) increases slightly in the three-shot setting. 
Llama similarly achieves its highest BERTScore with one-shot prompting (0.761), followed by small declines with additional demonstrations. These results suggest that, for the evaluated models and task, introducing a small amount of in-context guidance can be beneficial, whereas increasing the demonstration count does not necessarily yield further gains. This highlights the importance of considering both model-specific responsiveness and prompt efficiency when selecting the number of demonstrations.

\subsection{Effect of Prompting Strategies}
Motivated by the preceding context and ICL analyses, we next examine the effect of prompting strategies using the $paragraph_{1 \rightarrow 3}$ context setting. For this comparison, we retain three-shot prompting as a fixed common demonstration setting across models and prompt variants to ensure a controlled comparison, rather than because three-shot prompting is optimal for every model.
In addition to Gemini and Llama, we include GPT-4o as a widely used proprietary multilingual LLM to provide a broader comparison across model families. Table~\ref{tab:result_BNaP_vs_XLP} reports the performance of four prompting templates, \textit{BNaP+Baseline}, \textit{BNaP+MultiGen}, \textit{XLP+Baseline}, and \textit{XLP+MultiGen}, allowing us to analyze the combined effects of prompt language and contextual enrichment on Bengali headline generation.

\begin{table}[t!]
\centering
\caption{Performance comparison of prompting paradigms and contextual enrichment variants across different LLMs in a three-shot setting with partial article input ($paragraph_{1 \rightarrow 3}$).}
\label{tab:result_BNaP_vs_XLP}
\resizebox{\linewidth}{!}{%
\begin{tabular}{c c l c c c}
\toprule
Paradigm & Variant & Models & BERTScore \uparr & ROUGE-L \uparr & METEOR \uparr\\
\midrule

\multirow{12}{*}{\textit{BNaP}} 
  & \multirow{5}{*}{\texttt{Baseline}} 
    & Llama-3.3 & \makecell[t]{$0.754_{\color{ash}\pm 0.062}$\\{\color{ash}\small[0.749, 0.760]}} & \makecell[t]{$0.250_{\color{ash}\pm 0.195}$\\{\color{ash}\small[0.234, 0.268]}} & \makecell[t]{$0.164_{\color{ash}\pm 0.187}$\\{\color{ash}\small[0.149, 0.181]}}\\
  &  & {GPT-4o} & \makecell[t]{$0.764_{\color{ash}\pm 0.067}$\\{\color{ash}\small[0.758, 0.770]}} & \makecell[t]{$0.275_{\color{ash}\pm 0.208}$\\{\color{ash}\small[0.258, 0.293]}} & \makecell[t]{$0.231_{\color{ash}\pm 0.243}$\\{\color{ash}\small[0.210, 0.253]}}\\
  &  & Gemini-2.0 & \makecell[t]{$0.760_{\color{ash}\pm 0.069}$\\{\color{ash}\small[0.754, 0.766]}} & \makecell[t]{$0.283_{\color{ash}\pm 0.210}$\\{\color{ash}\small[0.266, 0.302]}} & \makecell[t]{$0.251_{\color{ash}\pm 0.246}$\\{\color{ash}\small[0.230, 0.273]}}\\
\cmidrule{2-6}
  & \multirow{5}{*}{\texttt{MultiGen}} 
    & Llama-3.3 & \makecell[t]{$0.746_{\color{ash}\pm 0.068}$\\{\color{ash}\small[0.740, 0.751]}} & \makecell[t]{$0.222_{\color{ash}\pm 0.198}$\\{\color{ash}\small[0.205, 0.239]}} & \makecell[t]{$0.146_{\color{ash}\pm 0.191}$\\{\color{ash}\small[0.130, 0.163]}}\\
  &  & GPT-4o & \makecell[t]{$0.759_{\color{ash}\pm 0.062}$\\{\color{ash}\small[0.753, 0.764]}} & \makecell[t]{$0.253_{\color{ash}\pm 0.189}$\\{\color{ash}\small[0.237, 0.270]}} & \makecell[t]{$0.194_{\color{ash}\pm 0.212}$\\{\color{ash}\small[0.176, 0.213]}}\\
  &  & Gemini-2.0 & \makecell[t]{$0.758_{\color{ash}\pm 0.066}$\\{\color{ash}\small[0.753, 0.764]}} & \makecell[t]{$0.273_{\color{ash}\pm 0.206}$\\{\color{ash}\small[0.256, 0.292]}} & \makecell[t]{$0.230_{\color{ash}\pm 0.236}$\\{\color{ash}\small[0.211, 0.252]}}\\

\midrule

\multirow{12}{*}{\textit{XLP}} 
  & \multirow{5}{*}{\texttt{Baseline}} 
    & Llama-3.3 & \makecell[t]{$0.758_{\color{ash}\pm 0.064}$\\{\color{ash}\small[0.752, 0.764]}} & \makecell[t]{$0.253_{\color{ash}\pm 0.197}$\\{\color{ash}\small[0.236, 0.271]}} & \makecell[t]{$0.173_{\color{ash}\pm 0.198}$\\{\color{ash}\small[0.156, 0.191]}}\\
  &  & GPT-4o & \makecell[t]{$0.758_{\color{ash}\pm 0.065}$\\{\color{ash}\small[0.752, 0.763]}} & \makecell[t]{$0.267_{\color{ash}\pm 0.200}$\\{\color{ash}\small[0.251, 0.284]}} & \makecell[t]{$0.240_{\color{ash}\pm 0.243}$\\{\color{ash}\small[0.220, 0.261]}}\\
  &  & Gemini-2.0 & \makecell[t]{$0.767_{\color{ash}\pm 0.073}$\\{\color{ash}\small[0.760, 0.773]}} & \makecell[t]{$0.301_{\color{ash}\pm 0.222}$\\{\color{ash}\small[0.282, 0.321]}} & \makecell[t]{$0.256_{\color{ash}\pm 0.254}$\\{\color{ash}\small[0.234, 0.279]}}\\
\cmidrule{2-6}
  & \multirow{5}{*}{\texttt{MultiGen}} 
    & Llama-3.3 & \makecell[t]{$0.757_{\color{ash}\pm 0.064}$\\{\color{ash}\small[0.752, 0.763]}} & \makecell[t]{$0.247_{\color{ash}\pm 0.200}$\\{\color{ash}\small[0.230, 0.265]}} & \makecell[t]{$0.166_{\color{ash}\pm 0.198}$\\{\color{ash}\small[0.149, 0.184]}}\\
  &  & GPT-4o & \makecell[t]{$0.753_{\color{ash}\pm 0.060}$\\{\color{ash}\small[0.748, 0.758]}} & \makecell[t]{$0.243_{\color{ash}\pm 0.184}$\\{\color{ash}\small[0.227, 0.259]}} & \makecell[t]{$0.202_{\color{ash}\pm 0.221}$\\{\color{ash}\small[0.183, 0.222]}}\\
  &  & Gemini-2.0 & \makecell[t]{$\textbf{0.769}_{\color{ash}\pm 0.072}$\\{\color{ash}\small[0.763, 0.776]}} & \makecell[t]{$0.292_{\color{ash}\pm 0.228}$\\{\color{ash}\small[0.273, 0.313]}} & \makecell[t]{$0.244_{\color{ash}\pm 0.256}$\\{\color{ash}\small[0.223, 0.268]}} \\

\bottomrule
\end{tabular}
}
\end{table}

\subsubsection{Prompt Language -- \textit{BNaP} vs. \textit{XLP}}
The effect of prompt language is model-dependent, although \textit{XLP} generally provides advantages for Llama and Gemini. As shown in Table~\ref{tab:result_BNaP_vs_XLP}, both models consistently achieve higher \texttt{BERTScore}s under \textit{XLP} than \textit{BNaP} across both prompting variants. For example, Llama improves from 0.754 (95\% CI: 0.749--0.760) under \textit{BNaP+Baseline} to 0.758 (95\% CI: 0.752--0.764) under \textit{XLP+Baseline},
while Gemini increases from 0.760 (95\% CI: 0.754--0.766) to 0.767 (95\% CI: 0.760--0.773).
Although these improvements do not reach statistical significance, they exhibit a consistent positive trend.

The largest gains are observed under the \textit{MultiGen} configuration, where \textit{XLP} improves \texttt{BERTScore} by 1.5\% for both Llama ($\Delta$ \texttt{BERTScore} = 0.012, 95\% CI: 0.0036--0.02; $p=0.0038$) and Gemini ($\Delta$ \texttt{BERTScore} = 0.011, 95\% CI: 0.004--0.018; $p=0.003$) relative to \textit{BNaP}. These statistically significant improvements indicate that cross-lingual prompting is particularly effective when combined with contextual enrichment.

In contrast, GPT-4o exhibits the opposite trend, performing slightly better under \textit{BNaP} than \textit{XLP}. Taken together, the results suggest that the effectiveness of prompt language depends on model-specific multilingual capabilities. For Llama and Gemini, English-language instructions appear to provide more effective task guidance, potentially enabling better utilization of auxiliary contextual cues and multilingual representations acquired during large-scale pre-training.

\subsubsection{Contextual Enrichment -- \textit{Baseline} vs. \textit{MultiGen}}
The effectiveness of contextual enrichment depends strongly on the prompting paradigm. Under \textit{BNaP}, the simpler \textit{Baseline} configuration consistently outperforms \textit{MultiGen} across all three models. 
For example, Gemini declines from 0.760 (95\% CI: 0.754--0.766) under \textit{BNaP+Baseline} to 0.758 (95\% CI: 0.753--0.764) under \textit{BNaP+MultiGen}.
Although the difference is not statistically significant ($\Delta$ \texttt{BERTScore} = $-0.0014$, 95\% CI: $-0.008$ to 0.006; $p=0.7$), the degradation is consistent across all models, suggesting that additional contextual cues do not provide substantial benefits when instructions are written entirely in Bengali.

In contrast, \textit{XLP} demonstrates greater compatibility with contextual enrichment. For Gemini, \textit{MultiGen} improves \texttt{BERTScore} from 0.767 (95\% CI: 0.760--0.773) to 0.769 (95\% CI: 0.763--0.776), yielding the strongest overall performance observed in the study. This improvement is statistically significant, although modest in magnitude ($\Delta$ \texttt{BERTScore} = 0.0025, 95\% CI: 0.0023--0.0074, $p = 0.0148$), suggesting that contextual enrichment provides a measurable but limited benefit for Gemini when combined with cross-lingual prompting.
Llama and GPT-4o, however, experience a modest decline. 
Overall, these results indicate that the effectiveness of auxiliary contextual cues is model-dependent: contextual enrichment provides a measurable but limited benefit for Gemini under XLP, while Llama and GPT-4o show no similar gains.

\subsection{Comparison with Fine-Tuned Models}
We compare our prompting-based in-context learning approach with the fine-tuned transformer models reported in \textit{BeliN}~\cite{OSAMA2025100138}. Prior work evaluated multilingual encoder--decoder models, including \textit{mT5}, \textit{mBART}, and \textit{BanglaT5}, using supervised fine-tuning and contextual enrichment through the \textit{MultiGen} framework. In contrast, our approach leverages general-purpose LLMs without task-specific training, relying solely on prompt design and in-context examples for adaptation.

\begin{table}[!t]
\centering
\caption{Comparison between previously reported fine-tuned transformer models and the LLM-based prompting approach used in this study.}
\label{tab:comparison}
\resizebox{\linewidth}{!}{%
\begin{tabular}{l l lll}
\toprule
Study & Model & BERTScore \uparr & ROUGE-L \uparr & METEOR \uparr \\

\midrule
$\text{\citet{OSAMA2025100138}}^\dagger$ 
    & mT5        & 0.717 & 0.169 & 0.109 \\
    & mT0       & 0.726 & 0.215 & 0.144 \\
    & BanglaT5   & \underline{0.751} & \underline{0.242} & \underline{0.167} \\ 
    & mBART     & 0.746 & 0.226 & 0.146 \\
    \hline
This study 
    &  Gemini-2.0*  & \textbf{0.769} (\textcolor{green!60!black}{+0.018}) & \textbf{0.292} (\textcolor{green!60!black}{+0.05}) & \textbf{0.244} (\textcolor{green!60!black}{+0.077}) \\
\bottomrule
\multicolumn{4}{l}{$^\dagger$Scores are reproduced from the cited study}\\
\multicolumn{4}{l}{*\textit{XLP+MultiGen $\cdot$ 3-shot}}
\end{tabular}%
}
\end{table}

As shown in Table~\ref{tab:comparison}, Gemini using the \textit{XLP+MultiGen} prompting configuration with three-shot prompting achieves higher scores than the best-performing fine-tuned \textit{BanglaT5} model across all evaluation metrics. Specifically, \texttt{BERTScore} improves from 0.751 to 0.769 (a relative improvement of 2.4\%), \texttt{ROUGE-L} from 0.242 to 0.292, and \texttt{METEOR} from 0.167 to 0.244. Notably, these gains are obtained without task-specific fine-tuning, suggesting that carefully designed prompting strategies and in-context learning can enable modern LLMs to achieve competitive performance and, in this benchmark, exceed the reported fine-tuned models on automatic metrics.
However, this comparison should be interpreted in light of differences in model scale, pre-training data, and adaptation mechanisms between fine-tuned transformers and large proprietary LLMs.

\begin{table}[!t]
\centering
\caption{Comparison between the best-performing fine-tuned model reported in prior work and the Gemini-based prompting approach used in this study on the \textit{Shironaam} benchmark~\cite{akash-etal-2023-shironaam}.}
\label{tab:external_evaluation}
\resizebox{\linewidth}{!}{%
\begin{tabular}{@{}l llll@{}}
\toprule
Model & BERTScore $\uparrow$ & ROUGE-L $\uparrow$ & METEOR $\uparrow$ & BLEU $\uparrow$ \\ \midrule
$\text{BERT2BERT \cite{akash-etal-2023-shironaam}}^\dagger$  & \textbf{0.831} & \textbf{0.503} & \textbf{0.435} & 0.318 \\
Gemini-2.0* (this study) & $0.810_{\color{ash}\pm 0.081}$ & $0.448_{\color{ash}\pm 0.251}$ & $0.397_{\color{ash}\pm 0.293}$ & $\textbf{0.373}_{\color{ash}\pm 0.248}$ \\
 & \color{ash}\small[0.808, 0.811] & \color{ash}\small[0.444, 0.452] & \color{ash}\small[0.393, 0.402] & \color{ash}\small[0.370, 0.378]\\
\bottomrule
\multicolumn{4}{l}{$^\dagger$Scores are reproduced from the cited study}\\
\multicolumn{4}{l}{*\textit{XLP+MultiGen $\cdot$ 3-shot}}
\end{tabular}%
}
\end{table}

\subsection{Evaluation on a Second Benchmark}
To assess the transferability of the selected prompting configuration beyond \textit{BeliN}, we evaluate it on the \textit{Shironaam} test dataset (articles $n=15,012$) and compare it with the best-performing fine-tuned \textit{BERT2BERT} model reported in prior work~\cite{akash-etal-2023-shironaam}. Unlike the supervised fine-tuning setup used in \textit{Shironaam}, our approach employs Gemini under a three-shot \textit{XLP+MultiGen} prompting configuration without task-specific parameter updates. Because the auxiliary information available in Shironaam differs from BeliN, we adapt the MultiGen prompt to the dataset-provided contextual cues, using category, topic words, and image-caption information.
The exact \textit{Shironaam}-specific \textit{XLP+MultiGen} template is provided in Table~\ref{tab:prompt_templates_xlp_multigen_shironaam} (Appendix~\ref{Prompt_templates}).

As shown in Table~\ref{tab:external_evaluation}, the fine-tuned \textit{BERT2BERT} model achieves higher scores on \texttt{BERTScore}, \texttt{ROUGE-L}, and \texttt{METEOR}, indicating stronger semantic alignment and structural correspondence with the reference headlines. Specifically, Gemini obtains a \texttt{BERTScore} of 0.810 compared with 0.831 for \textit{BERT2BERT}, while its \texttt{ROUGE-L} and \texttt{METEOR} scores are also lower. This advantage likely reflects the benefits of supervised task-specific optimization and direct exposure to labeled training data. However, Gemini achieves a higher \texttt{BLEU} score than \textit{BERT2BERT} (0.373 vs. 0.318), suggesting that in-context learning with a large multilingual LLM can produce lexically competitive headlines even without explicit fine-tuning.

Overall, the results highlight the complementary strengths of supervised fine-tuning and prompting-based adaptation. Fine-tuned encoder--decoder models remain more effective for maximizing task-specific performance on this benchmark, whereas selective context conditioning combined with in-context learning provides a flexible alternative that achieves competitive performance without additional model training. These findings further suggest that carefully designed prompting strategies can be applied across datasets, although supervised fine-tuning remains stronger on several metrics.

\subsection{Limitations}
This study has several limitations.  
First, our analysis is limited to a small set of proprietary and open-source LLMs, and the observed trends may vary across other multilingual models and prompting configurations. In addition, the effectiveness of prompting strategies is sensitive to prompt formulation, demonstration selection, demonstration ordering, and random variation across repeated runs, which may affect reproducibility and generalizability across datasets and tasks. Future work should therefore examine alternative few-shot demonstration strategies, such as random, similarity-based, and diversity-based selection, as well as prompt-order and repeated-run ablations.

Second, our evaluation relies primarily on automatic metrics without large-scale human assessment of factual consistency, fluency, or editorial quality. Future work may therefore explore adaptive context selection, automated prompt optimization, and human-centered evaluation for multilingual headline generation. Extending the analysis to larger and more diverse datasets, as well as multimodal and retrieval-augmented settings, would further strengthen the understanding of selective context conditioning in document-level generation tasks.

\section{Conclusion}
In this study, we investigated Bengali news headline generation using LLMs through selective context conditioning and prompting strategies. Our findings show that increasing input context does not necessarily improve generation quality; instead, conditioning on selected lead paragraphs of articles can maintain comparable performance and, for some models, improve over full-document conditioning. Through semantic alignment analysis, we demonstrate that the initial paragraphs of news articles contain most headline-relevant information, providing empirical support for relevance-focused context selection.

Our experiments further reveal that prompting strategies substantially influence the quality of the generated news headlines. Cross-Lingual Prompting (\textit{XLP}) often improves performance for Gemini and Llama, but its benefits are model-dependent, while Bengali Native Prompting (\textit{BNaP}) benefits from concise and minimally structured prompts. 
We also find that the benefit of in-context learning is strongly model-dependent. For Gemini, introducing a single demonstration captures most of the observed improvement, with little additional gain from two or three demonstrations, while Llama achieves its highest performance in the one-shot setting and does not benefit consistently from additional examples. These findings suggest that increasing the number of demonstrations is not necessarily advantageous for Bengali headline generation.

Overall, this work highlights the importance of contextual salience, prompt design, and model-specific behavior in multilingual document-level generation tasks. The findings suggest that carefully designed prompting and selective context conditioning can enable LLMs to achieve competitive headline generation performance for low-resource languages without task-specific fine-tuning.



\appendix

\section{Headline-Paragraph Semantic Alignment}\label{append:a}
Algorithm~\ref{alg:headline_similarity} is used only for the training-set semantic-alignment analysis that characterizes the positional distribution of headline-relevant information. It is not used to select paragraphs during test-time headline generation. The resulting training-set distribution motivates the position-based context settings evaluated subsequently, where the first $k$ article paragraphs are selected independently of the test reference headline.
\begin{algorithm}[h]
\caption{Identifying the paragraph most semantically aligned with the reference headline}
\label{alg:headline_similarity}

\KwIn{Dataset $D=\{(A_i,h_i)\}_{i=1}^{N}$ of news articles $A_i$ and their reference headlines $h_i$; sentence-embedding model $M$}
\KwOut{Most-aligned paragraph indices $I_{\max}$, paragraphs $P_{\max}$, and similarity scores $S_{\max}$}

$I_{\max}\leftarrow[]$; 
$P_{\max}\leftarrow[]$; 
$S_{\max}\leftarrow[]$\;

\For{each article--headline pair $(A,h)\in D$}{
    $P=\{p_1,p_2,\ldots,p_m\}\leftarrow \texttt{get\_paragraphs}(A)$\;
    $e_h\leftarrow M.\texttt{encode}(h)$\;
    $S\leftarrow[]$\;

    \For{$j\leftarrow1$ \KwTo $m$}{
        $e_j\leftarrow M.\texttt{encode}(p_j)$\;
        $s_j\leftarrow \mathrm{CosineSimilarity}(e_j,e_h)$\;
        $S.\mathrm{append}(s_j)$\;
    }

    $j^{*}\leftarrow \operatorname*{arg\,max}_{j} S[j]$\;
    $p^{*}\leftarrow p_{j^{*}}$\;
    $s^{*}\leftarrow S[j^{*}]$\;

    $I_{\max}.\mathrm{append}(j^{*})$\;
    $P_{\max}.\mathrm{append}(p^{*})$\;
    $S_{\max}.\mathrm{append}(s^{*})$\;
}

\Return{$I_{\max}, P_{\max}, S_{\max}$}\;
\end{algorithm}

\vspace{-1em}
\section{Prompt Templates}
\label{Prompt_templates}


\captionof{table}{Prompt template for \textit{XLP+Baseline}.}
\label{tab:prompt_templates_xlp_baseline}
\vspace{-1em}
\begin{tcolorbox}
\#\#\# Instruction: Generate a news headline in Bengali for the given news article.  

\# News Article: \{article\}  

\# News Headline: 
\end{tcolorbox}

\captionof{table}{Prompt template for \textit{BNaP+Baseline}.}
\label{tab:prompt_templates_bnap_baseline}
\vspace{-1em}
\begin{tcolorbox}
\bangla{\#\#\# নির্দেশনা: প্রদত্ত সংবাদ প্রতিবেদনের জন্য একটি বাংলা শিরোনাম তৈরি করুন।}  

\bangla{\# সংবাদ প্রতিবেদন:} \{article\}  

\bangla{\# সংবাদের শিরোনাম:}  
\end{tcolorbox}

\captionof{table}{\textit{XLP+MultiGen+1-shot} prompt template incorporating contextual cues from the \textit{Shironaam} corpus.}
\label{tab:prompt_templates_xlp_multigen_shironaam}
\begin{tcolorbox}
\#\#\# \textbf{Instruction:} Generate a news headline for the given news article considering the category, topic words and image caption of the news. Detailed information on these inputs is given below.

\textbf{News Category:} The broad section or domain of the news (e.g., Politics, Sports, Religion, Technology).

\textbf{News Topic words:} Important keywords or topic words that highlight the main subjects, entities, or themes discussed in the article.

\textbf{News Image Caption:} A brief description of the image associated with the news article.
\vspace{1em}

\# Example-1:

\# News Category: \{example1\_news\_category\}

\# News Topic words: \{example1\_news\_tags\}

\# News Image Caption: \{example1\_news\_image\_caption\}

\# News Article: \{example1\_news\_article\}

\# News Headline: \{example1\_news\_headline\}












\vspace{1em}

\# News Category: \{category\}

\# News Topic words: \{tags\}

\# News Image Caption: \{caption\}

\# News Article: \{article\}

\# News Headline:
\end{tcolorbox}

\captionof{table}{\textit{BNaP+MultiGen} prompt template incorporating contextual cues from the \textit{BeliN} corpus.}
\label{tab:prompt_templates_bnap_multigen}
\begin{tcolorbox}
\bangla{\#\#\# নির্দেশনা: প্রদত্ত সংবাদ প্রতিবেদনের জন্য সংবাদের অনুভূতি, সংবাদের বিভাগ এবং সংবাদের দৃষ্টিভঙ্গির ভিত্তিতে একটি বাংলা শিরোনাম তৈরি করুন। এই ইনপুটগুলির বিস্তারিত তথ্য নীচে দেওয়া হল।}  

\bangla{সংবাদের অনুভূতি: সংবাদের অনুভূতি ইতিবাচক, নেতিবাচক বা নিরপেক্ষ হতে পারে। একটি ইতিবাচক অনুভূতি সমর্থন নির্দেশ করে, একটি নেতিবাচক অনুভূতি সমালোচনা প্রকাশ করে এবং একটি নিরপেক্ষ অনুভূতি নিরপেক্ষতা বজায় রাখে।}  

\bangla{সংবাদের বিভাগ: এটি সংবাদের ধর্মীয় সংশ্লিষ্টতা নির্দেশ করে, যা ইসলাম, হিন্দুধর্ম, খ্রিস্টধর্ম, বৌদ্ধধর্ম বা অন্যান্য হতে পারে।}  

\bangla{সংবাদের দৃষ্টিভঙ্গি: এটি হলো সংবাদের প্রাথমিক ফোকাস বা থিম, যেমন ধর্মীয় প্রতিবেদন, উৎসব, শিক্ষা বা সংস্কৃতি।}  

\vspace{1em}

\bangla{\# সংবাদের অনুভূতি:} \{sentiment\} 

\bangla{\# সংবাদের বিভাগ:} \{category\}

\bangla{\# সংবাদের দৃষ্টিভঙ্গি:} \{aspect\} 

\bangla{\# সংবাদ প্রতিবেদন:} \{article\} 

\bangla{\# সংবাদের শিরোনাম:}  
\end{tcolorbox}

\vspace{3em}
\captionof{table}{\textit{XLP+MultiGen} prompt template incorporating contextual cues from the \textit{BeliN} corpus.}
\label{tab:prompt_templates_xlp_multigen}
\begin{tcolorbox}
\#\#\# \textbf{Instruction:} Generate a news headline for the given news article considering the sentiment of the news, the category of the news, and the perspective of the news. Detailed information on these inputs is given below.  

\textbf{News Sentiment:} The sentiment of the article can be positive, negative, or neutral. A positive sentiment indicates support, a negative sentiment conveys criticism, and a neutral sentiment maintains impartiality.  

\textbf{News Category:} The religious affiliation of the article, which can be Islam, Hinduism, Christianity, Buddhism, or Others.  

\textbf{News Aspect:} The primary focus or theme of the article, such as religious reports, festivals, education, or culture.  

\vspace{1em}

\# News Sentiment: \{sentiment\}  

\# News Category: \{category\}  

\# News Aspect: \{aspect\}  



\# News Article: \{article\}  

\# News Headline:  
\end{tcolorbox}

\section*{Ethics and Privacy Statement}
This work studies Bengali news headline generation using existing benchmark datasets and large language models. The study does not involve human-subject experiments, user interaction, or the collection of new personal data. The datasets used in this work consist of news articles and associated headline-generation metadata from previously released Bengali news benchmarks. We do not attempt to identify individuals, infer sensitive personal attributes, or generate user-specific profiles.

Nevertheless, automatic headline generation can have ethical implications. Generated headlines may be factually inaccurate, overly simplified, sensationalized, or inconsistent with the source article. Such errors may mislead readers, especially in news domains involving religion, politics, health, public safety, or other socially sensitive topics. Since one of the datasets focuses on religious news, biased or inaccurate headlines could also amplify social or cultural sensitivities if deployed without oversight.

The proposed study is intended for research evaluation rather than direct newsroom deployment. To mitigate potential harms, we evaluate generated headlines against reference headlines using established automatic metrics and explicitly acknowledge the limitation that automatic metrics do not fully capture factuality, fairness, fluency, or editorial appropriateness. In practical use, LLM-generated headlines should be reviewed by human editors before publication, particularly for sensitive topics. Future work should include human evaluation of factual consistency, bias, and editorial quality.

There are also privacy considerations when using LLMs for news generation. Although the benchmark data are drawn from news articles rather than private user records, news text may contain names, locations, organizations, or other personally identifying information already present in the source articles. Our experiments do not introduce additional personal data, and model outputs are generated only from the provided article context. However, any real-world deployment should ensure compliance with dataset licenses, data-use conditions, and applicable privacy regulations, especially when processing non-public or user-provided documents.

Finally, the technology could be misused to generate misleading, clickbait-style, or politically manipulative headlines at scale. We therefore recommend that headline-generation systems include provenance tracking, editorial review, factuality checks, and safeguards against deceptive or harmful use.


\bibliographystyle{ACM-Reference-Format}
\bibliography{custom}

@inproceedings{fan2018controllable,
    title = "Controllable Abstractive Summarization",
    author = "Fan, Angela  and
      Grangier, David  and
      Auli, Michael",
    editor = "Birch, Alexandra  and
      Finch, Andrew  and
      Luong, Thang  and
      Neubig, Graham  and
      Oda, Yusuke",
    booktitle = "Proceedings of the 2nd Workshop on Neural Machine Translation and Generation",
    month = {Jul},
    year = "2018",
    address = "Melbourne, Australia",
    publisher = "Association for Computational Linguistics",
    doi = "10.18653/v1/W18-2706",
    pages = "45--54",
}

@inproceedings{ravaut2024context,
    title = "On Context Utilization in Summarization with Large Language Models",
    author = "Ravaut, Mathieu  and
      Sun, Aixin  and
      Chen, Nancy  and
      Joty, Shafiq",
    editor = "Ku, Lun-Wei  and
      Martins, Andre  and
      Srikumar, Vivek",
    booktitle = "Proceedings of the 62nd Annual Meeting of the Association for Computational Linguistics (Volume 1: Long Papers)",
    month = aug,
    year = "2024",
    address = "Bangkok, Thailand",
    publisher = "Association for Computational Linguistics",
    doi = "10.18653/v1/2024.acl-long.153",
    pages = "2764--2781",
}

@inproceedings{litvak-etal-2019-ranlp,
    title = "{RANLP} 2019 Multilingual Headline Generation Task Overview",
    author = "Litvak, Marina  and
      Conroy, John M.  and
      Rankel, Peter A.",
    editor = "Giannakopoulos, George",
    booktitle = "Proceedings of the Workshop MultiLing 2019: Summarization Across Languages, Genres and Sources",
    month = sep,
    year = "2019",
    address = "Varna, Bulgaria",
    publisher = "INCOMA Ltd.",
    url = "https://aclanthology.org/W19-8901/",
    doi = "10.26615/978-954-452-058-8_001",
    pages = "1--5",
}

@inproceedings{bhattacharjee-etal-2023-banglanlg,
    title = "{B}angla{NLG} and {B}angla{T}5: Benchmarks and Resources for Evaluating Low-Resource Natural Language Generation in {B}angla",
    author = "Bhattacharjee, Abhik  and
      Hasan, Tahmid  and
      Ahmad, Wasi Uddin  and
      Shahriyar, Rifat",
    editor = "Vlachos, Andreas  and
      Augenstein, Isabelle",
    booktitle = "Findings of the Association for Computational Linguistics: EACL 2023",
    month = may,
    year = "2023",
    address = "Dubrovnik, Croatia",
    publisher = "Association for Computational Linguistics",
    url = "https://aclanthology.org/2023.findings-eacl.54/",
    doi = "10.18653/v1/2023.findings-eacl.54",
    pages = "726--735",
}

@inproceedings{li-etal-2023-crosslingual,
    title = "Crosslingual Retrieval Augmented In-context Learning for {B}angla",
    author = "Li, Xiaoqian  and
      Nie, Ercong  and
      Liang, Sheng",
    editor = "Alam, Firoj  and
      Kar, Sudipta  and
      Chowdhury, Shammur Absar  and
      Sadeque, Farig  and
      Amin, Ruhul",
    booktitle = "Proceedings of the First Workshop on Bangla Language Processing (BLP-2023)",
    month = dec,
    year = "2023",
    address = "Singapore",
    publisher = "Association for Computational Linguistics",
    url = "https://aclanthology.org/2023.banglalp-1.15/",
    doi = "10.18653/v1/2023.banglalp-1.15",
    pages = "136--151",
}

@inproceedings{chhabra-etal-2024-revisiting,
    title = "Revisiting Zero-Shot Abstractive Summarization in the Era of Large Language Models from the Perspective of Position Bias",
    author = "Chhabra, Anshuman  and
      Askari, Hadi  and
      Mohapatra, Prasant",
    editor = "Duh, Kevin  and
      Gomez, Helena  and
      Bethard, Steven",
    booktitle = "Proceedings of the 2024 Conference of the North American Chapter of the Association for Computational Linguistics: Human Language Technologies (Volume 2: Short Papers)",
    month = jun,
    year = "2024",
    address = "Mexico City, Mexico",
    publisher = "Association for Computational Linguistics",
    url = "https://aclanthology.org/2024.naacl-short.1/",
    doi = "10.18653/v1/2024.naacl-short.1",
    pages = "1--11",
}

@misc{googledeepmind2025gemini20flash,
  author       = {{Google DeepMind}},
  title        = {{Gemini 2.0 Flash: Model Card}},
  year         = {2025},
  month        = apr,
  howpublished = {Google DeepMind},
  url          = {https://storage.googleapis.com/deepmind-media/Model-Cards/Gemini-2-0-Flash-Model-Card.pdf},
  note         = {Published April 15, 2025}
}

@misc{meta2024llama33,
  author       = {{Meta}},
  title        = {{Llama 3.3 70B Instruct}},
  year         = {2024},
  howpublished = {Hugging Face Model Card},
  url          = {https://huggingface.co/meta-llama/Llama-3.3-70B-Instruct},
  note         = {Released December 6, 2024}
}

@misc{openai2024gpt4o,
  title         = {{GPT-4o System Card}},
  author        = {{OpenAI}},
  year          = {2024},
  eprint        = {2410.21276},
  archivePrefix = {arXiv},
  primaryClass  = {cs.CL},
  doi           = {10.48550/arXiv.2410.21276}
}

@inproceedings{banerjee2005meteor,
    title = "{METEOR}: An Automatic Metric for {MT} Evaluation with Improved Correlation with Human Judgments",
    author = "Banerjee, Satanjeev  and
      Lavie, Alon",
    editor = "Goldstein, Jade  and
      Lavie, Alon  and
      Lin, Chin-Yew  and
      Voss, Clare",
    booktitle = "Proceedings of the {ACL} Workshop on Intrinsic and Extrinsic Evaluation Measures for Machine Translation and/or Summarization",
    month = jun,
    year = "2005",
    address = "Ann Arbor, Michigan",
    publisher = "Association for Computational Linguistics",
    url = "https://aclanthology.org/W05-0909/",
    pages = "65--72"
}

@inproceedings{ahmed-etal-2025-bennumeval,
    title = "{B}en{N}um{E}val: A Benchmark to Assess {LLM}s' Numerical Reasoning Capabilities in {B}engali",
    author = "Ahmed, Kawsar  and
      Osama, Md  and
      Sharif, Omar  and
      Hossain, Eftekhar  and
      Hoque, Mohammed Moshiul",
    editor = "Che, Wanxiang  and
      Nabende, Joyce  and
      Shutova, Ekaterina  and
      Pilehvar, Mohammad Taher",
    booktitle = "Findings of the Association for Computational Linguistics: ACL 2025",
    month = jul,
    year = "2025",
    address = "Vienna, Austria",
    publisher = "Association for Computational Linguistics",
    url = "https://aclanthology.org/2025.findings-acl.915/",
    doi = "10.18653/v1/2025.findings-acl.915",
    pages = "17782--17799",
    ISBN = "979-8-89176-256-5"
}

@inproceedings{ding2023harnessing,
    title = "Harnessing the power of {LLM}s: Evaluating human-{AI} text co-creation through the lens of news headline generation",
    author = "Ding, Zijian  and
      Smith-Renner, Alison  and
      Zhang, Wenjuan  and
      Tetreault, Joel  and
      Jaimes, Alejandro",
    editor = "Bouamor, Houda  and
      Pino, Juan  and
      Bali, Kalika",
    booktitle = "Findings of the Association for Computational Linguistics: EMNLP 2023",
    month = dec,
    year = "2023",
    address = "Singapore",
    publisher = "Association for Computational Linguistics",
    url = "https://aclanthology.org/2023.findings-emnlp.217/",
    doi = "10.18653/v1/2023.findings-emnlp.217",
    pages = "3321--3339",
    
}

@inproceedings{lewis2019bart,
    title = "{BART}: Denoising Sequence-to-Sequence Pre-training for Natural Language Generation, Translation, and Comprehension",
    author = "Lewis, Mike  and
      Liu, Yinhan  and
      Goyal, Naman  and
      Ghazvininejad, Marjan  and
      Mohamed, Abdelrahman  and
      Levy, Omer  and
      Stoyanov, Veselin  and
      Zettlemoyer, Luke",
    editor = "Jurafsky, Dan  and
      Chai, Joyce  and
      Schluter, Natalie  and
      Tetreault, Joel",
    booktitle = "Proceedings of the 58th Annual Meeting of the Association for Computational Linguistics",
    month = jul,
    year = "2020",
    address = "Online",
    publisher = "Association for Computational Linguistics",
    url = "https://aclanthology.org/2020.acl-main.703/",
    doi = "10.18653/v1/2020.acl-main.703",
    pages = "7871--7880"
}

@article{raffel2020exploring,
  title={Exploring the limits of transfer learning with a unified text-to-text transformer},
  author={Raffel, Colin and Shazeer, Noam and Roberts, Adam and Lee, Katherine and Narang, Sharan and Matena, Michael and Zhou, Yanqi and Li, Wei and Liu, Peter J},
  journal={Journal of machine learning research},
  volume={21},
  number={140},
  pages={1--67},
  year={2020}
}

@inproceedings{jiang-etal-2024-instruction,
    title = "Instruction-tuned Language Models are Better Knowledge Learners",
    author = "Jiang, Zhengbao  and
      Sun, Zhiqing  and
      Shi, Weijia  and
      Rodriguez, Pedro  and
      Zhou, Chunting  and
      Neubig, Graham  and
      Lin, Xi Victoria  and
      Yih, Wen-tau  and
      Iyer, Srinivasan",
    editor = "Ku, Lun-Wei  and
      Martins, Andre  and
      Srikumar, Vivek",
    booktitle = "Proceedings of the 62nd Annual Meeting of the Association for Computational Linguistics (Volume 1: Long Papers)",
    month = aug,
    year = "2024",
    address = "Bangkok, Thailand",
    publisher = "Association for Computational Linguistics",
    url = "https://aclanthology.org/2024.acl-long.296/",
    doi = "10.18653/v1/2024.acl-long.296",
    pages = "5421--5434",
}

@inproceedings{kojima2022large,
 author = {Kojima, Takeshi and Gu, Shixiang (Shane) and Reid, Machel and Matsuo, Yutaka and Iwasawa, Yusuke},
 booktitle = {Advances in Neural Information Processing Systems},
 doi = {10.52202/068431-1613},
 editor = {S. Koyejo and S. Mohamed and A. Agarwal and D. Belgrave and K. Cho and A. Oh},
 pages = {22199--22213},
 publisher = {Curran Associates, Inc.},
 title = {Large Language Models are Zero-Shot Reasoners},
 url = {https://proceedings.neurips.cc/paper_files/paper/2022/file/8bb0d291acd4acf06ef112099c16f326-Paper-Conference.pdf},
 volume = {35},
 address = "online",
 year = {2022}
}

@article{liu2023lost,
    title = "Lost in the Middle: How Language Models Use Long Contexts",
    author = "Liu, Nelson F.  and
      Lin, Kevin  and
      Hewitt, John  and
      Paranjape, Ashwin  and
      Bevilacqua, Michele  and
      Petroni, Fabio  and
      Liang, Percy",
    journal = "Transactions of the Association for Computational Linguistics",
    volume = "12",
    year = "2024",
    address = "Cambridge, MA",
    publisher = "MIT Press",
    url = "https://aclanthology.org/2024.tacl-1.9/",
    doi = "10.1162/tacl_a_00638",
    pages = "157--173"
}

@inproceedings{hasan2021xl,
    title = "{XL}-Sum: Large-Scale Multilingual Abstractive Summarization for 44 Languages",
    author = "Hasan, Tahmid  and
      Bhattacharjee, Abhik  and
      Islam, Md. Saiful  and
      Mubasshir, Kazi  and
      Li, Yuan-Fang  and
      Kang, Yong-Bin  and
      Rahman, M. Sohel  and
      Shahriyar, Rifat",
    editor = "Zong, Chengqing  and
      Xia, Fei  and
      Li, Wenjie  and
      Navigli, Roberto",
    booktitle = "Findings of the Association for Computational Linguistics: ACL-IJCNLP 2021",
    month = aug,
    year = "2021",
    address = "Online",
    publisher = "Association for Computational Linguistics",
    url = "https://aclanthology.org/2021.findings-acl.413/",
    doi = "10.18653/v1/2021.findings-acl.413",
    pages = "4693--4703"
}

@article{OSAMA2025100138,
title = {BeliN: A novel corpus for Bengali religious news headline generation using contextual feature fusion},
journal = {Natural Language Processing Journal},
volume = {11},
pages = {100138},
year = {2025},
issn = {2949-7191},
doi = {https://doi.org/10.1016/j.nlp.2025.100138},
author = {Md Osama and Ashim Dey and Kawsar Ahmed and Muhammad Ashad Kabir},
}

@misc{zhang2020bertscoreevaluatingtextgeneration,
      title={BERTScore: Evaluating Text Generation with BERT}, 
      author={Tianyi Zhang and Varsha Kishore and Felix Wu and Kilian Q. Weinberger and Yoav Artzi},
      year={2020},
      eprint={1904.09675},
      archivePrefix={arXiv},
      primaryClass={cs.CL},
      url={https://arxiv.org/abs/1904.09675}, 
}

@inproceedings{lin2004rouge,
    title = "{ROUGE}: A Package for Automatic Evaluation of Summaries",
    author = "Lin, Chin-Yew",
    booktitle = "Text Summarization Branches Out",
    month = jul,
    year = "2004",
    address = "Barcelona, Spain",
    publisher = "Association for Computational Linguistics",
    url = "https://aclanthology.org/W04-1013/",
    pages = "74--81"
}

@article{beltagy2020longformer,
  title={Longformer: The Long-Document Transformer},
  author={Iz Beltagy and Matthew E. Peters and Arman Cohan},
  journal={ArXiv},
  year={2020},
  pages = {},
  volume={abs/2004.05150},
  url={https://api.semanticscholar.org/CorpusID:215737171}
}

@inproceedings{qin2023cross,
    title = "Cross-lingual Prompting: Improving Zero-shot Chain-of-Thought Reasoning across Languages",
    author = "Qin, Libo  and
      Chen, Qiguang  and
      Wei, Fuxuan  and
      Huang, Shijue  and
      Che, Wanxiang",
    editor = "Bouamor, Houda  and
      Pino, Juan  and
      Bali, Kalika",
    booktitle = "Proceedings of the 2023 Conference on Empirical Methods in Natural Language Processing",
    month = dec,
    year = "2023",
    address = "Singapore",
    publisher = "Association for Computational Linguistics",
    url = "https://aclanthology.org/2023.emnlp-main.163/",
    doi = "10.18653/v1/2023.emnlp-main.163",
    pages = "2695--2709",
}

@inproceedings{akash-etal-2023-shironaam,
    title = "Shironaam: {B}engali News Headline Generation using Auxiliary Information",
    author = "Akash, Abu Ubaida  and
      Nayeem, Mir Tafseer  and
      Shohan, Faisal Tareque  and
      Islam, Tanvir",
    editor = "Vlachos, Andreas  and
      Augenstein, Isabelle",
    booktitle = "Proceedings of the 17th Conference of the European Chapter of the Association for Computational Linguistics",
    month = may,
    year = "2023",
    address = "Dubrovnik, Croatia",
    publisher = "Association for Computational Linguistics",
    url = "https://aclanthology.org/2023.eacl-main.4/",
    doi = "10.18653/v1/2023.eacl-main.4",
    pages = "52--67",
}

@article{po2003news,
author = {Po¨ttker, Horst},
year = {2003},
month = {11},
pages = {501-511},
title = {News and its communicative quality: the inverted pyramid—when and why did it appear?},
volume = {4},
journal = {Journalism Studies},
doi = {10.1080/1461670032000136596}
}

\end{document}